\documentclass[10pt,twocolumn,letterpaper]{article}

\usepackage{cvpr}
\usepackage{times}
\usepackage{epsfig}
\usepackage{graphicx}
\usepackage{amsmath}
\usepackage{amssymb}
\usepackage{float}

\usepackage[breaklinks=true,bookmarks=false]{hyperref}

\cvprfinalcopy 

\def\cvprPaperID{****} 

\begin{document}

\title{Dual-stream Maximum Self-attention Multi-instance Learning}

\author{Bin Li\\
University of Wisconsin at Madison\\
Wisconsin, WI 53705\\
{\tt\small bli346@wisc.edu}
\and
Kevin W. Eliceiri\\
University of Wisconsin at Madison\\
Wisconsin, WI 53705\\
{\tt\small eliceiri@wisc.edu}
}

\maketitle

\begin{abstract}
Multi-instance learning (MIL) is a form of weakly supervised learning where a single class label is assigned to a bag of instances while the instance-level labels are not available. Training classifiers to accurately determine the bag label and instance labels is a challenging but critical task in many practical scenarios, such as computational histopathology. 
Recently, MIL models fully parameterized by neural networks have become popular due to the high flexibility and superior performance. Most of these models rely on attention mechanisms that assign attention scores across the instance embeddings in a bag and produce the bag embedding using an aggregation operator. 
In this paper, we proposed a dual-stream maximum self-attention MIL model (DSMIL) parameterized by neural networks. The first stream deploys a simple MIL max-pooling while the top-activated instance embedding is determined and used to obtain self-attention scores across instance embeddings in the second stream. 
Different from most of the previous methods, the proposed model jointly learns an instance classifier and a bag classifier based on the same instance embeddings. The experiments results show that our method achieves superior performance compared to the best MIL methods and demonstrates state-of-the-art performance on benchmark MIL datasets.
\end{abstract}

\section{Introduction}

Typical supervised learning problems assume that each training sample in the dataset has a label, and the classifier can be trained by using the label as instance level supervise signal. However, in many practical scenarios, getting instance level labels can be difficult due to the high complexity and intensive labor for labeling each individual instance. 
In these cases, a label is assigned to a group of instances instead. This problem is called multi-instance learning (MIL) \cite{dietterich_solving_1997}. 
For example, in whole slide image (WSI) \cite{ghaznavi_digital_2013, higgins_applications_2015} analysis, the images can have tremendously large dimensions but usually the whole image is assigned with a single label while region-level annotation is seldom given \cite{campanella_clinical-grade_2019}.

Recently, Ilse et al. \cite{ilse_attention-based_2018} introduced the Attention-based MIL model fully parameterized by neural networks. The aggregation operator as well as the feature extractor are end-to-end trainable and can aggregation instance embeddings to a bag embedding. The attention-mechanism used in the model assigns an attention score to each of the instance embeddings, and the final classifier operates on a bag embedding which is a gated-sum of the instance embeddings. The attention score reflect how much an instance is likely to be the key instance that trigger the bag classifier. Later on, \cite{rymarczyk_kernel_2020, wang_hyperspectral_2019, miyazaki_weakly-supervised_2020} proposed to use self-attention mechanism \cite{vaswani_attention_2017} to further consider the the dependencies between instance embeddings. However, computing the self-attention matrix across all instance embeddings in a bag is computationally complex and might yield redundant information that does not contributes useful supervising signal. More importantly, both of the MIL models can have difficulties to solve clinical WSI image classification problems in practical scenarios where the WSIs produce tens of thousands of patches. The memory requirement for training a deep CNN-based feature extractor as well as the following aggregation operator requires gradients to flow through the CNN of all patches, which prohibits the training of the bag embedding-based model.

In this paper, we show that the cross matching of all queries in self-attention for MIL is sub-optimal, and a matching using only the top-activated queries does not only reduce the computational complexity but also improves the classification performance. 
we propose a novel dual-stream MIL (DSMIL) model parameterized by neural networks that jointly learns an instance classifier and a bag classifier. 
The first stream of the model deploys a standard MIL max pooling, which determines the top-activated embeddings.
In the second stream, the attention score is computed across the instances by correlating only the top-activated queries with the instances in the bag. 
Compute the proposed maximum self-attention requires a computation complexity of $\mathrm{O}(n)$ compared to a full self-attention between each pair of instance embeddings which is $\mathrm{O}(n^2)$, and distributes clearer supervised signal to each of the instances. 
The max self-attention mechanism applies a soft selection of the instances by comparing how similar each instance is to the top-activated instance. 
After training, the first stream can be solely used on individual instance as an instance level classifier, which can directly detects the key instances inside the bag and does not require obtaining the attention scores in the scope of the whole bag. The training of the model can alter between the two streams to allow the training of the feature extractor as well as the following aggregation operator when the bag size is huge. In the experiments, we show that our model outperforms other best MIL models by a large margin on MIL benchmark datasets.

\section{Method}
\subsection{Problem formulation}
In MIL, a group of training samples is considered as a \textit{bag}, a bag is a collection of \textit{instances} where number of instances can vary. 
Each bag has a bag class label, it is positive if the bag contains at least one instance of that class and it is negative if there is no such instance in the bag. Meanwhile, there is no instance-level label available. 
In the case of binary classification, let $B={(x_1, y_1),...,(x_n,y_n)}$ be a bag where $x_i \in \chi$ are instances with labels $y_i \in {0, 1}$, the label of $B$ is given by 
\begin{equation}
    c(B)=1-\prod_{i=1}^{n} (1-y_i)
\label{eq:bag_1}
\end{equation}
Suppose there are some suitable transformations $f$ and $g$, such that
\begin{equation}
    c(B)=g(f(x_0),...f(x_n))
\end{equation}
The multi-instance learning problem can thus be modeled in two ways based on the functions of $f$ and $g$: 1) $f$ is a instance-level classifier that produces a class score for each instance, $g$ is a pooling operator that aggregate the instance scores to produce a bag score. 2) $f$ is a instance-level feature extractor that maps each instance to a embedding, $g$ is an aggregation function that first maps all instance embeddings to a bag embedding and produces a bag score based on the bag embedding. The first way is called instance-based method while the latter one is called embedding-based method. Under the multi-instance learning assumption $g$ needs to be permutation-invariant in either case. 

It is shown that the embedding-approach generally performs better in terms of the bag level classification accuracy \cite{wang_revisiting_2018}. The instance-level classifier in the instance-based method can be trained insufficiently and yield noisy results, due to that the instance-level labels are unknown and the supervised signal comes from a handcrafted pooling operator such as max pooling or average pooling. The embedding-based method produces a bag score based on a bag representation which is directly supervised by the bag label and usually yields better accuracy compared to the instance-based method, however, it is hard to determine the key instances that trigger the classifier, in contrast to the instance-based method \cite{liu_detecting_2017}. 
\begin{figure*}[!ht]
    \centering
    \includegraphics[width=\textwidth]{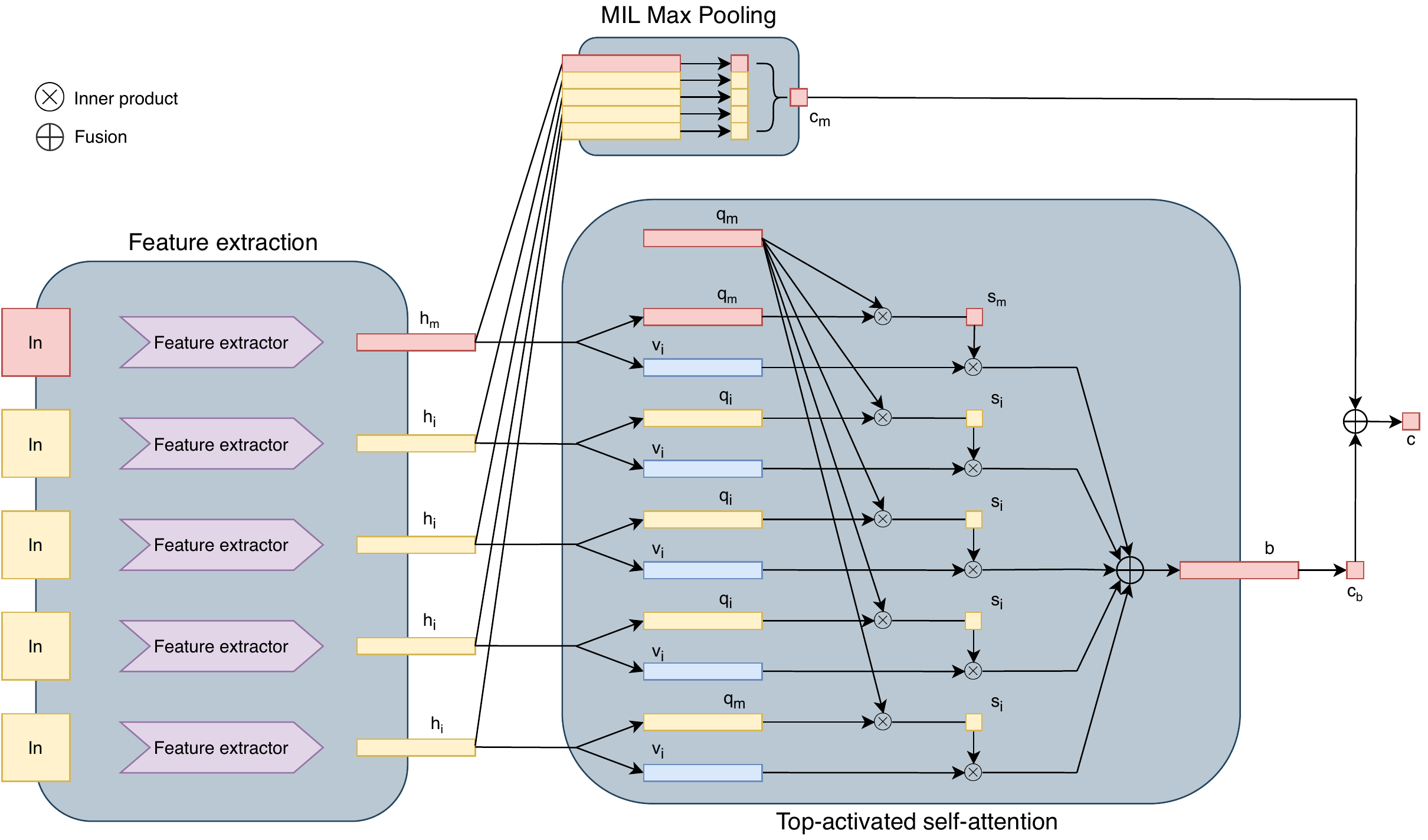}
    \caption{Dual-stream MIL aggregation with max self-attention mechanism.}
    \label{fig:model}
\end{figure*}
\subsection{Dual-stream MIL aggregation}
In contrast to most previous methods, we proposed a MIL model, DSMIL, that jointly learns an instance classifier and a bag classifier with a dual-stream architecture where the supervised signal flows between two streams. 
Let $\mathbf{H} = [ \mathbf{h}_0,...,\mathbf{h}_{N-1} ] \in \mathbb{R}^{L \times N}$ be a bag of instance embeddings where $\mathbf{h}_i\in\mathbb{R}^{L \times 1}$ is the embedding of $i$th instance. The first stream is an instance level classifier with MIL max pooling, which is given by
\begin{equation}
    \mathrm{\mathbf{c}}_m = 
    \mathrm{max}
    \{ \mathrm{\mathbf{W}}_0 \mathrm{\mathbf{h}}_0, \dots, \mathrm{\mathbf{W}}_{0} \mathrm{\mathbf{h}}_{N-1}\}
\end{equation}
where $\mathbf{W}_0$ is the weight matrix of a fully connected layer.

The second stream learns a bag embedding from instance embeddings and a bag classifier that scores the bag embedding. We obtain the top-activation instance embedding $\mathbf{h}_{m}$ from the first stream and transform each instance embedding $\mathbf{h}_{i}$ into two feature vectors, queries $\mathbf{q}_{i}\in \mathbb{R}^{L \times 1}$ and information $\mathbf{v}_{i}\in \mathbb{R}^{L \times 1}$, which are given respectively by
\begin{equation}
    \mathbf{q}_{i} = \mathbf{W}_q \mathbf{h}_{i}, \quad \mathbf{v}_{i} = \mathbf{W}_v \mathbf{h}_{i}, \quad
    i=0, \dots, N-1
\end{equation}
Where $\mathbf{W}_q$ and $\mathbf{W}_v$ are weight matrices of two fully connected layers. Each element of the max self-attention vector is then given by
\begin{equation}
    a_i = 
    \frac{\mathrm{exp}(\mathrm{s}_{i})}
    {\sum_{i=0}^{N-1} \mathrm{exp}(\mathrm{s}_{i})}, \quad 
    s_i= \langle \mathbf{q}_i, \mathbf{q}_m \rangle, \quad
    i=0, \dots, N-1
\end{equation}
Instead of matching each query with additional key vectors, the query is matched with other queries and no key vectors are learned. The softmax layer ensures the weights summed to 1 regardless of the bag size. The bag embedding $\mathbf{b} \in \mathbb{R}^{L \times 1}$ is then given by:
\begin{equation}
    \mathrm{\mathbf{b}}= \sum_{i} a_i \mathbf{v}_{i}
\end{equation}
The sum operator denotes element-wise sum.

The bag score $\mathbf{c} \in \mathbb{R}^{C \times 1}$ is then given by:
\begin{equation}
    \mathrm{\mathbf{c}}_b = 
    \mathrm{\mathbf{W}}_1
    \mathrm{\mathbf{b}}
\end{equation}
Where $\mathbf{W}_1$ is a weight matrix of a fully connected layer.
The final score of the bag is the weighted sum of the scores of the two streams
\begin{equation}
    \mathrm{\mathbf{\hat{c}}} = (1-\lambda) \mathrm{\mathbf{c}}_m + \lambda \mathrm{\mathbf{c}}_b,
    \quad \lambda \in [0, 1]
\end{equation}
The proposed self-attention mechanism allows information to be extracted element-wisely from each instance embedding according to their similarity to the top-activated embedding and produces a bag embedding with a constant shape regardless of the bag size.
The model is illustrated in \textbf{Fig. \ref{fig:model}}. Noted that both the max pooling stream and the aggregation stream uses the same instance embeddings, and the first-stage feature extractor is thus trained using supervision signal from both streams. As a result, it can be trained much sufficiently compared to the case where there is only the max pooling stream. After the model is trained, the max pooling stream of DSMIL can be taken as a standalone instance classifier with much better performance compared to the one train solely with MIL max pooling.

\section{Experiments and results}
\subsection{MIL datasets without feature extractor}
We carried out benchmark experiments on several MIL benchmark datasets. The first group of datasets consist of feature vectors of the instances and does not require a feature extractor to be learned in the first stage. First two datasets (MUSK1, MUSK2) are used to predict drug effect based on the molecule conformations. Same molecules can have different conformations and only some of them may be effective conformations. 
Thus, bags containing molecules with their different conformations are made \cite{dietterich_solving_1997}. For each bag, there are multiple conformations of the same molecule and the bag is labeled positive if at least one conformation is effective, negative otherwise. 
The other three datasets, ELEPHANT, FOX and TIGER, consists of feature vectors extracted from images. Each bag is made up of a group of segments of an image and the bag is labeled as positive if at least one segment contains the animal of interest, negative if there is no such animal presented. Details regarding the datasets can be found in the Apendix of \cite{ilse_attention-based_2018}. 

Since the feature vectors are already given, the implementation involves directly feeding the feature vectors to the dual-stream aggregation operator. The first stream scores each instance using a fully connected layer and max pooling over all instance scores. The top-activated instance embedding is determined and passed to the self-attention stream for computing self-attention matrix described above. The aggregation operator is implemented in the same way in the following experiments. 
Experiments were run 5 times each with a 10-fold cross-validation. The mean and standard deviation of the classification accuracy is reported in \textbf{Tab. \ref{tab:classic}}. The benchmark results show that DSMIL outperforms the best MIL models by a large margin and demonstrates state-of-the-art performance on these benchmark datasets.

\begin{table*}
    \centering
    \begin{tabular}{c|c|c|c|c|c}
    \hline
         Methods & MUSK1 & MUSK2 & FOX & TIGER & ELEPHANT \\
         \hline
         mi-SVM & 0.874 $\pm$ N/A & 0.836 $\pm$ N/A & 0.582 $\pm$ N/A & 0.784 $\pm$ N/A &  0.822 $\pm$ N/A \\
         MI-SVM & 0.779 $\pm$ N/A & 0.843 $\pm$ N/A & 0.578 $\pm$ N/A & 0.840 $\pm$ N/A &  0.843 $\pm$ N/A \\
         MI-Kernel & 0.880 $\pm$ 0.031 & 0.893 $\pm$ 0.015 & 0.603 $\pm$ 0.028 & 0.842 $\pm$ 0.010 &  0.843 $\pm$ 0.016 \\
         EM-DD & 0.849 $\pm$ 0.044 & 0.869 $\pm$ 0.048 & 0.609 $\pm$ 0.045 & 0.730 $\pm$ 0.043 &  0.771 $\pm$ 0.043 \\
         mi-Graph & 0.889 $\pm$ 0.033 & 0.903 $\pm$ 0.039 & 0.620 $\pm$ 0.044 & 0.860 $\pm$ 0.037 &  0.869 $\pm$ 0.035 \\
         miVLAD & 0.871 $\pm$ 0.043 & 0.872 $\pm$ 0.042 & 0.620 $\pm$ 0.044 & 0.811 $\pm$ 0.039 &  0.850 $\pm$ 0.036 \\
         miFV & 0.909 $\pm$ 0.040 & 0.884 $\pm$ 0.042 & 0.621 $\pm$ 0.049 & 0.813 $\pm$ 0.037 &  0.852 $\pm$ 0.036 \\
         \hline
         mi-Net & 0.889 $\pm$ 0.039 & 0.858 $\pm$ 0.049 & 0.613 $\pm$ 0.035 & 0.824 $\pm$ 0.034 &  0.858 $\pm$ 0.037 \\
         MI-Net & 0.887 $\pm$ 0.041 & 0.859 $\pm$ 0.046 & 0.622 $\pm$ 0.038 & 0.830 $\pm$ 0.032 &  0.862 $\pm$ 0.034 \\
         MI-Net with DS & 0.894 $\pm$ 0.042 & 0.874 $\pm$ 0.043 & 0.630 $\pm$ 0.037 & 0.845 $\pm$ 0.039 &  0.872 $\pm$ 0.032 \\
         MI-Net with RC & 0.898 $\pm$ 0.043 & 0.873 $\pm$ 0.044 & 0.619 $\pm$ 0.047 & 0.836 $\pm$ 0.037 &  0.857 $\pm$ 0.040 \\
         AbMILP & 0.892 $\pm$ 0.040 & 0.858 $\pm$ 0.048 & 0.615 $\pm$ 0.043 & 0.839 $\pm$ 0.022 &  0.868 $\pm$ 0.022 \\
         AbMILP-Gated & 0.900 $\pm$ 0.050 & 0.863 $\pm$ 0.042 & 0.603 $\pm$ 0.029 & 0.845 $\pm$ 0.018 &  0.857 $\pm$ 0.027 \\
         GNN-MIL & 0.917 $\pm$ 0.048 & 0.892 $\pm$ 0.011 & 0.679 $\pm$ 0.007 & \textbf{0.876 $\pm$ 0.015} &  0.903 $\pm$ 0.010 \\
         \hline 
         DSMIL & \textbf{0.923 $\pm$ 0.019} & \textbf{0.914 $\pm$ 0.023} & \textbf{0.767 $\pm$ 0.015} & 0.871 $\pm$ 0.023 &  \textbf{0.907 $\pm$ 0.013} \\
         \hline
    \end{tabular}
    \caption{Table 1. Performance comparison on classical MIL dataset. Experiments were run 5 times each with a 10-fold cross-validation. The mean and standard deviation of the classification accuracy is reported (mean $\pm$ std). mi-SVM \cite{andrews_support_2003}, MI-SVM \cite{andrews_support_2003}, MI-Kernel \cite{gartner_multi-instance_2002}, EM-DD \cite{zhang_em-dd_2002}, mi-Graph \cite{zhou_multi-instance_2009} miVLAD \cite{wei_scalable_2017}, miFV \cite{wei_scalable_2017}, mi-Net\cite{wang_revisiting_2018}, MI-Net \cite{wang_revisiting_2018}, MI-Net with DS \cite{wang_revisiting_2018}, MI-Net with RC \cite{wang_revisiting_2018}, AbMILP \cite{ilse_attention-based_2018}, AbMILP-Gated \cite{ilse_attention-based_2018}, GNN-MIL \cite{tu_multiple_2019}. Previous benchmark results are taken from \cite{ilse_attention-based_2018}}
    \label{tab:classic}
\end{table*}

\begin{table*}[!ht]
    \centering
    \begin{tabular}{c|c|c|c}
    \hline
         Number of training bags & 50 & 100 & 300 \\
         \hline
         Instance+max & 0.553 $\pm$ 0.053 & 0.745 $\pm$ 0.100 & \textbf{0.984 $\pm$ 0.001} \\
         Instance+mean & 0.663 $\pm$ 0.014 & 0.676 $\pm$ 0.012 & 0.709 $\pm$ 0.02 \\
         MI+SVM[1] & 0.697 $\pm$ 0.054 & 0.851 $\pm$ 0.009 & 0.926 $\pm$ 0.004 \\
         Embedded+max & 0.713 $\pm$ 0.016 & 0.914 $\pm$ 0.011 & 0.980 $\pm$ 0.001 \\
         Embedded+mean & 0.695 $\pm$ 0.026 & 0.814 $\pm$ 0.027 & 0.974 $\pm$ 0.002 \\
         AbMILP[2] & 0.768 $\pm$ 0.054 & \textbf{0.948 $\pm$ 0.007} & 0.980 $\pm$ 0.001 \\
         AbMILP-Gated[2] & 0.753 $\pm$ 0.054 & 0.916 $\pm$ 0.013 & 0.980 $\pm$ 0.004 \\
         \hline
         DSMIL & \textbf{0.894 $\pm$ 0.017} & \textbf{0.948 $\pm$ 0.012} & 0.979 $\pm$ 0.001 \\
         \hline
    \end{tabular}
    \caption{Performance comparison on MNIST-Bag dataset with average 10 instances per bag (10 $\pm$ 2). Experiments were run 5 times. The mean and standard deviation of the AUC is reported (mean $\pm$ std). MI-SVM \cite{andrews_support_2003}, AbMILP \cite{ilse_attention-based_2018}, AbMILP-Gated \cite{ilse_attention-based_2018}.}
    \label{tab:minst10}
\end{table*}

\begin{table*}[!ht]
    \centering
    \begin{tabular}{c|c|c|c}
    \hline
         Number of training bags & 50 & 100 & 300 \\
         \hline
         Instance+max & 0.576 $\pm$ 0.059 & 0.715 $\pm$ 0.096 & 0.994 $\pm$ 0.001 \\
         Instance+mean & 0.737 $\pm$ 0.014 & 0.744 $\pm$ 0.029 & 0.722 $\pm$ 0.021 \\
         MI-SVM[1] & 0.824 $\pm$ 0.067 & 0.946 $\pm$ 0.004 & 0.975 $\pm$ 0.001 \\
         Embedded+max & 0.872 $\pm$ 0.039 & 0.984 $\pm$ 0.005 & \textbf{0.996 $\pm$ 0.001} \\
         Embedded+mean & 0.841 $\pm$ 0.013 & 0.906 $\pm$ 0.046 & \textbf{0.996 $\pm$ 0.001} \\
         AbMILP[2] & 0.967 $\pm$ 0.010 & 0.982 $\pm$ 0.003 & 0.989 $\pm$ 0.003 \\
         AbMILP-Gated[2] & 0.920 $\pm$ 0.042 & 0.977 $\pm$ 0.003 & 0.994 $\pm$ 0.002 \\
         \hline
         DSMIL & \textbf{0.975 $\pm$ 0.001} & \textbf{0.991 $\pm$ 0.001} & 0.993 $\pm$ 0.001 \\
         \hline
    \end{tabular}
    \caption{Performance comparison on MNIST-Bag dataset with average 50 instances per bag (50 $\pm$ 10). Experiments were run 5 times. The mean and standard deviation of the AUC is reported (mean $\pm$ std). MI-SVM \cite{andrews_support_2003}, AbMILP \cite{ilse_attention-based_2018}, AbMILP-Gated \cite{ilse_attention-based_2018}.}
    \label{tab:minst50}
\end{table*}

\begin{table*}[!ht]
    \centering
    \begin{tabular}{c|c|c|c}
    \hline
         Number of training bags & 50 & 100 & 300 \\
         \hline
         Instance+max & 0.543 $\pm$ 0.054 & 0.804 $\pm$ 0.107 & \textbf{1.000 $\pm$ 0.000} \\
         Instance+mean & 0.842 $\pm$ 0.023 & 0.855 $\pm$ 0.025 & 0.859 $\pm$ 0.029 \\
         MI-SVM[1] & 0.871 $\pm$ 0.060 & 0.991 $\pm$ 0.004 & 0.997 $\pm$ 0.001 \\
         Embedded+max & 0.977 $\pm$ 0.009 & 0.999 $\pm$ 0.001 & \textbf{1.000 $\pm$ 0.000} \\
         Embedded+mean & 0.959 $\pm$ 0.010 & 0.990 $\pm$ 0.003 & \textbf{1.000 $\pm$ 0.000} \\
         AbMILP[2] & 0.966 $\pm$ 0.001 & 0.998 $\pm$ 0.001 & \textbf{1.000 $\pm$ 0.000} \\
         AbMILP-Gated[2] & 0.998 $\pm$ 0.001 & 0.999 $\pm$ 0.000 & \textbf{1.000 $\pm$ 0.000} \\
         \hline
         DSMIL & \textbf{0.999 $\pm$ 0.000} & \textbf{1.000 $\pm$ 0.000} & \textbf{1.000 $\pm$ 0.000} \\
         \hline
    \end{tabular}
    \caption{Performance comparison on MNIST-Bag dataset with average 100 instances per bag (100 $\pm$ 20). Experiments were run 5 times. The mean and standard deviation of the AUC is reported (mean $\pm$ std). MI-SVM \cite{andrews_support_2003}, AbMILP \cite{ilse_attention-based_2018}, AbMILP-gated \cite{ilse_attention-based_2018}.}
    \label{tab:minst100}
\end{table*}

\begin{figure*}[!ht]
    \centering
    \includegraphics[width=\textwidth]{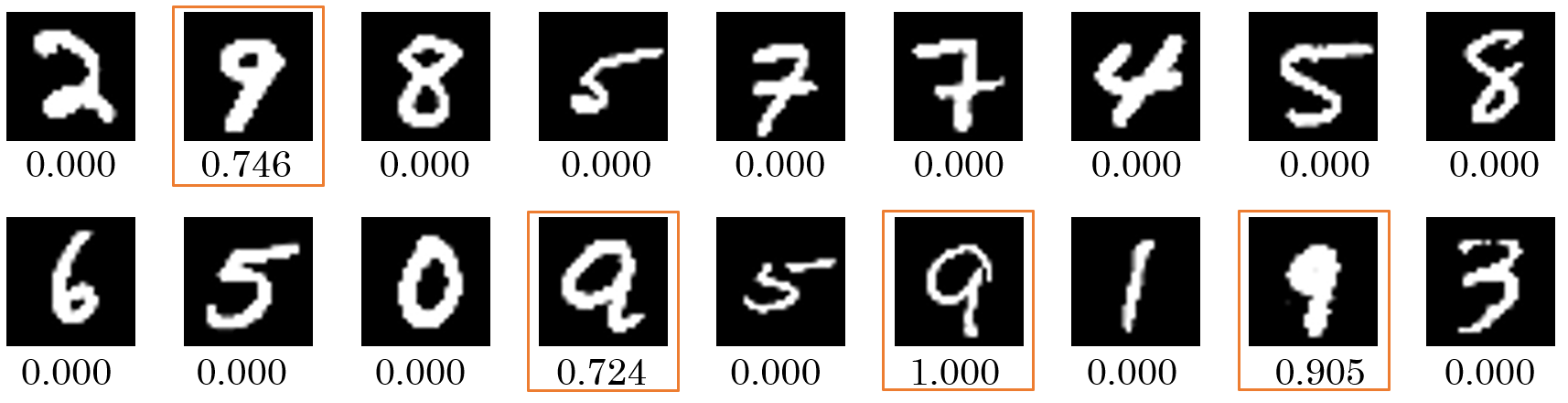}
    \caption{Classifier from the max pooling stream can be used as instance level classifier.}
    \label{fig:mnist}
\end{figure*}

We also compared the performance of DSMIL to the methods reported in \cite{ilse_attention-based_2018} on the MNIST-Bag dataset. The MNIST dataset is split into a fix division of a training set and a testing set. The bags are constructed to have multiple images from the MNIST dataset and the number '9' is chosen to be the positive class. The bag is labeled as positive if it contains at least one '9' and negative otherwise. We conducted experiments on settings of different number of training bags (50, 100, 300) under the conditions of different mean of the bag size (10, 50, 100). The number of testing bags is the same for all the experiments (1000 bags). In this experiment, the dataset is generated exactly same way as \cite{ilse_attention-based_2018}. Experiments were run 5 times, the mean and standard deviation of the AUC are reported in \textbf{Tab. \ref{tab:minst10}}, \textbf{Tab. \ref{tab:minst50}}, \textbf{Tab. \ref{tab:minst100}}. The benchmark results of other models are taken from \cite{ilse_attention-based_2018}. The proposed method outperforms the previous models by a large margin in the case where the number of training bags is small, and has comparable results when the number of training bags is large. We show that the instance classifier obtained from the max pooling stream can be directly used to score individual instances regardless of the bag size \textbf{Fig. \ref{fig:mnist}}.

\textit{Implementation details}. We used the LeNet5 \cite{lecun_gradient-based_1998} as the feature extractor, and passed the feature vectors to the proposed aggregation operator. The model is trained fully end-to-end by back-propagtion. We used a constant learning rate of 0.0001 to train the model 40 epochs in total with Adam optimizer.

\subsection{Histopathological dataset}
We evaluate the performance of proposed aggregation operator on two histopathological datasets that are used in several MIL studies. This first one is the breast cancer dataset \cite{drelie_gelasca_evaluation_2008} and the second one is a colon cancer dataset \cite{sirinukunwattana_locality_2016}.
\textbf{Breast cancer dataset}. This dataset contains 58 weakly labeled 896$\times$768 H\&E stained images. An image is labeled as malignant if at least one region contains breast cancer, and benign if it contains no cancer cells. Each image is divided into 32$\times$32 patches. A patch is discarded if the average saturation is below 0.05, which corresponds to the empty background. Each image roughly yields 600 to 700 valid patches.

\textbf{Colon cancer dataset}. This dataset contains 100 H\&E stained images. The images contain both normal and malignant regions. Several classes of nuclei (i.e. epithelial, inflammatory, fibroblast, and miscellaneous) are marked and in total consists of 22,444 marked nuclei. Each image is divided into 27$\times$27 patches which is treated as a bag. The weak labels are determined by whether the bag contains a least one marked epithelial cell which is most the place where colon cancer originates \cite{ricci-vitiani_identification_2007}. 

\begin{figure*}[!ht]
    \centering
    \includegraphics[width=\textwidth]{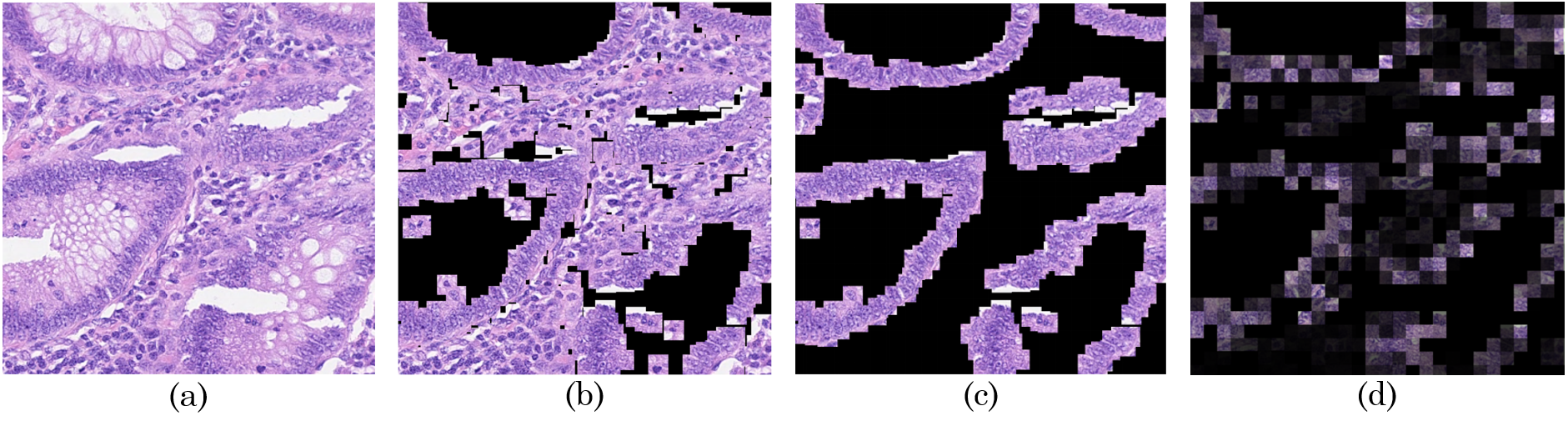}
    \caption{Visualization of the results on a representative image in the colon cancer dataset. (a) Input image. (b) All labelled cells. Patches are cropped centered to the coordinates list of the annotations. (c) Labelled epithelial cells. (d) Activation score of the instance classifier from the max pooling stream. Patches are extracted with 1/3 overlap and fused by averaging.}
    \label{fig:colon}
\end{figure*}

We evaluated the performance using metrics including accuracy, AUC, precision, recall, and F-score. High recall is especially important for the case of cancer detection since fail to detect the presence of cancer can lead to sever consequence. We ran the experiment 5 times each with a 10-fold cross-validation and summarized the results compared to reports in \cite{ilse_attention-based_2018} and \cite{rymarczyk_kernel_2020} in \textbf{Tab. \ref{tab:bcc}} and \textbf{Tab. \ref{tab:crc}}. DSMIL performs surprisingly well on the breast cancer dataset and outperforms other model by a significant margin. This might due to the difference in the data augmentation pipeline. We show that the max pooling stream can be used as a standalone instance classifier and perform epithelial cell detection in the colon dataset \textbf{Fig. \ref{fig:colon}}.

\begin{table*}[!ht]
    \centering
    \begin{tabular}{c|c|c|c|c|c}
    \hline
         Methods & Accuracy & Precision & Recall & F-score & AUC \\
         \hline
         AbMILP[1] & 0.904 $\pm$ 0.011 & \textbf{0.953 $\pm$ 0.014} & 0.855 $\pm$ 0.017 & 0.901 $\pm$ 0.011 & 0.968 $\pm$ 0.009 \\
         AbMILP-Gated[1] & 0.898 $\pm$ 0.020 & 0.944 $\pm$ 0.016 & 0.851 $\pm$ 0.035 & 0.893 $\pm$ 0.022 & 0.968 $\pm$ 0.010 \\
         SA-AbMILP[2] & 0.908 $\pm$ 0.013 & 0.938 $\pm$ 0.020 & 0.872 $\pm$ 0.024 & 0.890 $\pm$ 0.019 & 0.981 $\pm$ 0.070 \\
         GSA-AbMILP[2] & 0.884 $\pm$ 0.017 & 0.952 $\pm$ 0.017 & 0.837 $\pm$ 0.028 & 0.871 $\pm$ 0.022 & \textbf{0.985 $\pm$ 0.060} \\
         IQSA-AbMILP[2] & 0.890 $\pm$ 0.019 & 0.939 $\pm$ 0.021 & 0.855 $\pm$ 0.029 & 0.869 $\pm$ 0.025 & 0.966 $\pm$ 0.011 \\
         LSA-AbMILP[2]& 0.847 $\pm$ 0.017 & 0.927 $\pm$ 0.016 & 0.857 $\pm$ 0.027 & 0.874 $\pm$ 0.018 & 0.984 $\pm$ 0.050 \\
         \hline
         DSMIL & \textbf{0.911 $\pm$ 0.012} & 0.932 $\pm$ 0.013 & \textbf{0.912 $\pm$ 0.009} & \textbf{0.909 $\pm$ 0.04} & 0.967 $\pm$ 0.013\\
         \hline
    \end{tabular}
    \caption{Performance comparison on the colon cancer dataset. Experiments were run 5 times each with a 10-fold cross-validation. The mean and standard deviation of the accuracy, precision, reacal, F-score are reported. (mean $\pm$ std).}
    \label{tab:crc}
\end{table*}

\begin{table*}[!ht]
    \centering
    \begin{tabular}{c|c|c|c|c|c}
    \hline
         Methods & Accuracy & Precision & Recall & F-score & AUC \\
         \hline
         AbMILP[1] & 0.745 $\pm$ 0.018 & 0.718 $\pm$ 0.021 & 0.715 $\pm$ 0.046 & 0.712 $\pm$ 0.025 & 0.775 $\pm$ 0.016 \\
         AbMILP-Gated[1] & 0.755 $\pm$ 0.016 & 0.728 $\pm$ 0.016 & 0.731 $\pm$ 0.042 & 0.725 $\pm$ 0.023 & 0.799 $\pm$ 0.020 \\
         SA-AbMILP[2] & 0.750 $\pm$ 0.025 & 0.773 $\pm$ 0.037 & 0.749 $\pm$ 0.037 & 0.725 $\pm$ 0.025 & 0.859 $\pm$ 0.022 \\
         GSA-AbMILP[2] & 0.758 $\pm$ 0.021 & 0.793 $\pm$ 0.033 & 0.747 $\pm$ 0.034 & 0.725 $\pm$ 0.025 & 0.859 $\pm$ 0.022 \\
         IQSA-AbMILP[2] & 0.767 $\pm$ 0.022 & 0.786 $\pm$ 0.023 & 0.751 $\pm$ 0.042 & 0.667 $\pm$ 0.041 & 0.859 $\pm$ 0.021 \\
         LSA-AbMILP[2]& 0.655 $\pm$ 0.029 & 0.625 $\pm$ 0.037 & 0.895 $\pm$ 0.026 & 0.685 $\pm$ 0.026 & 0.867 $\pm$ 0.021 \\
         \hline
         DSMIL & \textbf{0.937 $\pm$ 0.00} & \textbf{0.925 $\pm$ 0.091} & \textbf{0.925 $\pm$ 0.003} & \textbf{0.974 $\pm$ 0.05} & \textbf{0.951 $\pm$ 0.013}\\
         \hline
    \end{tabular}
    \caption{Performance comparison on the breast cancer dataset. Experiments were run 5 times each with a 10-fold cross-validation. The mean and standard deviation of the accuracy, precision, reacal, F-score are reported. (mean $\pm$ std).}
    \label{tab:bcc}
\end{table*}

\textit{Implementation details}. We used the feature extractor as described in \cite{sirinukunwattana_locality_2016} with some minor modifications. Models are trained with Adam optimizer \cite{kingma_adam_2017} with learning rate of 0.0001. The model is trained 60 epochs and the learning rate is dropped by a factor of 0.1 at the 30th epoch. The loss is computed using mean square error which gives better gradient flow compared to negative log-likelihood. We implemented an online image augmentation pipeline involving rotation, flipping, affine distortion, aspect ratio distortion, and color jitter. 

\section{Conclusions and discussions}
In this paper, we proposed a dual-stream MIL (DSMIL) model that jointly learns an instance classifier and a bag classifier. The top-activated instance determined by the instance classifier is used to generate self-attention scores for each instance by correlating the top-activate query with all instance queries across the bag. The instance embeddings are then aggregated to a bag embedding according to the self-attention scores. This top-activated self-attention mechanism considers the dependencies of the instances to the top-activated instance and applies a soft selection on the instances based on their similarity to the top-activated query. Experiments show that this method does not only alleviate the computation complexity but also achieves superior performance on several benchmark datasets.

Moreover, the training of DSMIL can alter between the instance stream and the bag stream which the aggregation operator and the feature extractor it to be learned on extremely large bags. Since the max pooling stream and aggregation stream uses the same instance embeddings, the instance stream can be used standalone as an instance level classifier with the feature extractor being more sufficiently trained compared to the case of single stream MIL max pooling. The next step will be using this model to solve practical clinical problems such as whole slide image classification where the bag can contain up to tens of thousands of instances and may require much deeper CNNs as the feature extractor. The actual gain of performance on the instance classifier compared to single stream MIL max pooling is yet to be evaluated. Plus, the degree of degradation of performance when the training alternates between max pooling stream and aggregation stream is also yet to be seen. These are two important aspects for whole slide image classification, since when a deeper CNN is deployed as the feature extractor and the bag size becomes tremendously large, it is hard to feed both the whole aggregation operator and the CNN feature extractor into GPU for training.

{\small
\bibliographystyle{ieee}
\bibliography{references}
}

\end{document}